\documentclass[11pt]{article}

\usepackage{microtype} 
\usepackage{booktabs}  
\usepackage{url}  
\usepackage{graphicx}
\usepackage{amssymb}
\usepackage{algorithm}
\usepackage{algpseudocode}
\usepackage{wrapfig}
\usepackage{caption}
\usepackage{amsmath}
\usepackage{amsthm}
\usepackage[final,shortpaper]{automl}

\usepackage{natbib}
\title{Efficient Leakage-Free Neural Architecture Search under Leave-One-Subject-Out Evaluation}

\author[ ]{\nameemail{Heinke Hihn}{heinke.hihn@iu.org}}

\affil[ ]{IU International University of Applied Sciences, Berlin, Germany}

\hypersetup{%
  pdfauthor={}, 
  pdftitle={},
  pdfsubject={},
  pdfkeywords={}
}

\begin{document}

\maketitle

\begin{abstract}
Leave-One-Subject-Out (LOSO) evaluation estimates generalisation performance for subject-based classification but makes Neural Architecture Search (NAS) computationally expensive because a fully nested implementation requires $N$ independent architecture searches and, assuming approximately linear training cost, scales as $\mathcal{O}(N^2)$. We propose a leakage-free, block-based approach that shares NAS runs across subjects. On the BioVid Heat Pain dataset, our approach increased the mean accuracy from $82.79\%$ to $83.39\%$ while reducing the number of parameters by up to $99.2\%$.
\end{abstract}

\section{Method}
\begin{wrapfigure}{r}{0.5\textwidth}
    \centering
    \vspace{-\baselineskip}
    \begin{minipage}{\linewidth}
        \captionof{algorithm}{Block NAS with final LOSO evaluation}
        \label{alg:painnas-cross-fitted-loso}
        \small
        \begin{algorithmic}[1]
\Require Subjects $\mathcal S$, candidate architectures $\mathcal A$
\State Split $\mathcal S$ into disjoint outer blocks $\mathcal G_1,\ldots,\mathcal G_B$
\ForAll{outer blocks $\mathcal G_b$}
  \State Exclude $\mathcal G_b$ and fit each $a\in\mathcal A$ by rotating
    over subject-disjoint inner folds
  \State Compute the cross-validated metric $J_b(a)$ and select
    $a_b^\star\gets\arg\max_a J_b(a)$
  \ForAll{LOSO targets $j\in\mathcal G_b$}
    \State Fit $a_b^\star$ on all subjects $\mathcal S\setminus\{j\}$
    \State Evaluate once on held-out subject $j$
  \EndFor
\EndFor
\State \Return the mean held-out accuracy
\end{algorithmic}
    \end{minipage}
\end{wrapfigure}
Automated pain assessment aims to build systems classifying the pain level a subject experiences~\citep{khan2025systematic}. Although deep learning has become standard, architecture design has remained a manual task despite the potential benefits of Neural Architecture Search (NAS)~\citep{ren2021comprehensive}. We argue that the main reason is the computational cost of NAS in the LOSO setting. NAS has been applied to physiological signals
and multimodal fusion \citep{wu2023autoeer,duan2022cross,ghebriout2024harmonic}, but to our knowledge no previous method has addressed the repeated architecture searches required by subject-wise outer evaluation. 

Formally, the accuracy of Leave-One-Subject-Out cross-validation is $\operatorname{Acc}_{\mathrm{LOSO}}  =\frac{1}{N}\sum_{j=1}^{N}
\operatorname{Acc}\!\left(D_{-j},D_j\right)$, where \(\operatorname{Acc}\!\left(D_{-j},D_j\right)\) denotes the classification accuracy of a model trained on \(D_{-j}\), which excludes subject \(j\), and evaluated on that subject's data \(D_j\). Given \(N\) subjects, NAS and final model fitting are performed independently for each of the \(N\) folds using the remaining \(N-1\) subjects. To avoid data leakage, no samples of the held-out subject are used during optimisation. Thus, the total runtime is proportional to \(T_{\mathrm{LOSO\text{-}NAS}} = N\left[T_{\mathrm{NAS}}(N-1)+T_{\mathrm{fit}}(N-1)+T_{\mathrm{test}}(1)\right] \in \mathcal{O}(N^2),\) assuming \(T_{\mathrm{NAS}}(n),T_{\mathrm{fit}}(n)\in\mathcal{O}(n)\).
To reduce the number of NAS runs, we divide the subjects $\mathcal{S}$ into disjoint \textit{outer} blocks $\mathcal{G}_1,\ldots,\mathcal{G}_B$. For each outer block $b$, NAS uses only the training set $\mathcal{T}_b=\mathcal{S}\setminus\mathcal{G}_b$. To evaluate a proxy for LOSO generalisation, we further split $\mathcal{T}_b$ into $K$ subject-disjoint inner folds: For each candidate architecture $a$ and inner fold $k$, a model is initialised independently, fitted on the other \(K-1\) folds, and validated on fold $k$. The selection objective is given as 
\begin{equation}
\bar A_b(a)=\frac{1}{N_b}\sum_{i\in\mathcal{T}_b}A_i(a),
\qquad
\widehat{\sigma}_b(a)=
\sqrt{\frac{1}{N_b-1}\sum_{i\in\mathcal{T}_b}
\left(A_i(a)-\bar A_b(a)\right)^2},
\qquad
J_b(a)=\bar A_b(a)-\beta
\frac{\widehat{\sigma}_b(a)}{\sqrt{N_b}},
\label{eq:objective}
\end{equation}
where $\bar A_b(a)$ is the mean subject-level validation accuracy of architecture \(a\) over $\mathcal{T}_b$, $N_b=|\mathcal{T}_b|$ is the number of subjects in $\mathcal{T}_b$, $\widehat\sigma_b(a)$ is the sample standard deviation of the subject-level accuracies over $\mathcal{T}_b$, and $J_b(a)$ defines a selection score. It rewards mean subject-level accuracy and penalises estimates that exhibit high inter-subject variance controlled by a factor $\beta$. We used $\beta = 1.0$. Algorithm~\ref{alg:painnas-cross-fitted-loso} gives a high-level overview.

After the inner folds, we retain the checkpoint with the highest validation score of the architecture found by optimising Eq.~\ref{eq:objective}. The model is then trained for additional epochs using the samples of all 86 non-target subjects, including the former inner-validation subjects and the other members of $\mathcal{G}_b$ -- subjects in $\mathcal{G}_b\setminus\{j\}$
are included in the final fitting, whereas target subject $j$ remains completely excluded. Finally, it is evaluated once on the held-out data $D_j$. The final reported metric is the mean accuracy across all target-subjects. To implement NAS, we use an Optuna study~\citep{akiba2019optuna} with a tree-structured Parzen estimator on Eq.~\ref{eq:objective}. The implementation is made available at \href{https://github.com/hhihn/PainLOSONAS}{this repo}. 
\section{Results}
\begin{table*}[t]
    \centering
    \caption{%
    LOSO accuracies. Standard deviations are shown in parentheses where available.
    Bold indicates the best overall result, whereas underlining indicates
    the best result obtained using early fusion.
    }
    \label{tab:pain_recognition_overview}

    \small
    \setlength{\tabcolsep}{5pt}

    \begin{tabular*}{\textwidth}{
        @{\extracolsep{\fill}}
        lll
        lll
        @{}
    }
        \toprule
        \textbf{Method} & \textbf{Fusion} & \textbf{Accuracy}
        & \textbf{Method} & \textbf{Fusion} & \textbf{Accuracy} \\
        \midrule

        \cite{CrossModTransformer2025}
        & LF & \textbf{87.52\% ($\pm 11.0$)}
        &
        \cite{Steur2025Supervised}
        & LF & 84.22\% ($\pm 13.2$)
        \\

        \cite{thiam2025dealing}
        & LF & 85.32\% ($\pm 13.7$)
        &
        \cite{thiam2019exploring}
        & EF & 82.79\% ($\pm 15.2$)
        \\

        \cite{jiang2024personalized}
        & LF & 84.58\% ($\pm 13.3$)
        &
        \cite{werner2016automatic}
        & EF & 72.4\%
        \\

        \cite{thiam2019exploring}
        & LF & 84.40\% ($\pm 14.4$)
        &
        \cite{werner2014automatic}
        & EF & 77.8\%
        \\

        \cite{thiam2021multi}
        & LF & 84.20\% ($\pm 13.7$)
        &
        \cite{kachele2015multimodal}
        & EF & 78.9\%
        \\

        \midrule
        \multicolumn{4}{l}{
            Ours (block specific architectures)}
            & EF
            & \underline{83.39\%} ($\pm 14.4$) \\
        \bottomrule
    \end{tabular*}
\end{table*}

We evaluated on the BioVid heat pain dataset \citep{Walter2013BioVid} in the binary no-pain vs. pain setting and used the architecture proposed by \cite{thiam2019exploring} as a baseline. The 87 subjects were partitioned into five outer blocks and three inner folds. The outer block contained each $18$ or $17$ subjects. An $18$-subject block leaves $69$ training subjects, a $17$-subject block leaves $70$. Thus, five NAS studies replace 87 independent searches. We implemented the search on a Google Colab L4 22 GB GPU instance, where the runtime for a full 87-subject NAS, training, and evaluation was approximately 90h and reduced it to approximately 12h under the same configuration.

We use the early fusion (EF) approach in which the data is combined on a sensor (EDA, EMG, ECG) level, as this is efficient because it does not build separate modality streams unlike in late- or decision-fusion (LF) systems. LF models outperform EF~\cite{thiam2019exploring}, which is why current methods focus on LF architectures (e.g.,~\cite{khan2025systematic} report a single EF deep learning study). Our method increased the EF baseline from 82.79\% to 83.39\% while reducing the number of parameters by 76.7\% to 99.2\% to between $61.4\,\mathrm{k}$ and $1.79\,\mathrm{M}$ from $7.7\,\mathrm{M}$ -- see Table~\ref{tab:pain_recognition_overview} for the results.

\section{Discussion and Conclusion}
We have proposed a method to implement a computationally feasible NAS in the LOSO setting that improved the baseline and substantially reduced the number of parameters, increasing model efficiency. One limitation is the high variance of the architectural components and their parameter counts and how to mitigate this effect. Future research should investigate how to select an architecture for deployment, how the approach performs on the more difficult multiclass classification, whether it also improves LF architectures, and other classification problems under the LOSO setting such as affective computing~\citep{hihn2016inferring,hihn2016gestures}.
\bibliography{main}
\end{document}